\documentclass{article}

\usepackage{microtype}
\usepackage{graphicx}
\usepackage{subfigure}
\usepackage{booktabs} 
\usepackage{multirow}
\usepackage{array}

\usepackage{color}
\definecolor{hr}{gray}{0.63}  
\newcolumntype{*}{>{\global\let\currentrowstyle\relax}}
\newcolumntype{^}{>{\currentrowstyle}}
\newcommand{\rowstyle}[1]{\gdef\currentrowstyle{#1}#1\ignorespaces}

\usepackage{hyperref}

\usepackage[algo2e]{algorithm2e}
\usepackage{algorithmic}
\usepackage{algorithm}

 \usepackage[accepted]{icml2024}

\usepackage{amsmath}
\usepackage{pifont}
\usepackage{amsfonts}
\usepackage{amssymb}
\usepackage{mathtools}
\usepackage{amsthm}
\usepackage{bbm}

\usepackage[capitalize,noabbrev]{cleveref}

\theoremstyle{plain}

\theoremstyle{definition}

\theoremstyle{remark}

\usepackage[textsize=tiny]{todonotes}

\icmltitlerunning{Image Clustering with External Guidance}

\begin{document}

\twocolumn[
\icmltitle{Image Clustering with External Guidance}



\icmlsetsymbol{equal}{*}

\begin{icmlauthorlist}
\icmlauthor{Yunfan Li}{scu}
\icmlauthor{Peng Hu}{scu}
\icmlauthor{Dezhong Peng}{scu}
\icmlauthor{Jiancheng Lv}{scu}
\icmlauthor{Jianping Fan}{lenovo}
\icmlauthor{Xi Peng}{scu}
\end{icmlauthorlist}

\icmlaffiliation{scu}{School of Computer Science, Sichuan University, Chengdu, China}
\icmlaffiliation{lenovo}{AI Lab at Lenovo Research, Beijing, China}

\icmlcorrespondingauthor{Xi Peng}{pengx.gm@gmail.com}

\icmlkeywords{Clustering, Deep Clustering, External Knowledge}

\vskip 0.3in
]



\printAffiliationsAndNotice{}  

\begin{abstract}
The core of clustering lies in incorporating prior knowledge to construct supervision signals. From classic k-means based on data compactness to recent contrastive clustering guided by self-supervision, the evolution of clustering methods intrinsically corresponds to the progression of supervision signals. At present, substantial efforts have been devoted to mining internal supervision signals from data. Nevertheless, the abundant external knowledge such as semantic descriptions, which naturally conduces to clustering, is regrettably overlooked. In this work, we propose leveraging external knowledge as a new supervision signal to guide clustering. To implement and validate our idea, we design an externally guided clustering method (Text-Aided Clustering, TAC), which leverages the textual semantics of WordNet to facilitate image clustering. Specifically, TAC first selects and retrieves WordNet nouns that best distinguish images to enhance the feature discriminability. Then, TAC collaborates text and image modalities by mutually distilling cross-modal neighborhood information. Experiments demonstrate that TAC achieves state-of-the-art performance on five widely used and three more challenging image clustering benchmarks, including the full ImageNet-1K dataset. The code can be accessed at \url{https://github.com/XLearning-SCU/2024-ICML-TAC}.
\end{abstract}

\section{Introduction}
\label{introduction}

\begin{figure}[!t]
\centering
\includegraphics[width=\columnwidth]{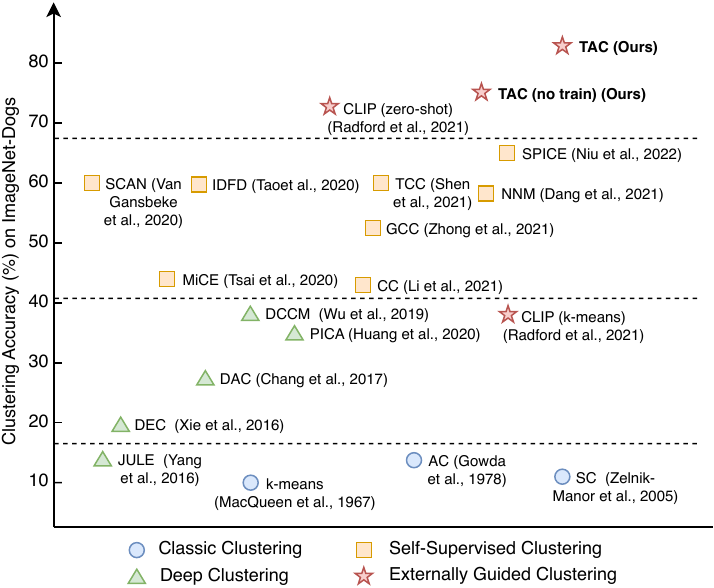}
\caption{The evolution of clustering methods could be roughly divided into three eras, including \textbf{i) classic clustering}, which designs clustering strategies based on data distribution assumptions; \textbf{ii) deep clustering}, which extracts clustering-favorable features with deep neural networks, and \textbf{iii) self-supervised clustering}, which constructs supervision signals through data augmentations or momentum strategies. In this work, instead of mining the internal supervision, we propose exploring external knowledge to facilitate image clustering. We categorize such a novel paradigm as \textbf{iv) externally guided clustering}. By leveraging the semantics in the text modality, our TAC pushes the clustering accuracy to a new state-of-the-art.}\label{fig: performance}
\end{figure}

Image clustering aims at partitioning images into different groups in an unsupervised fashion, which is a long-standing task in machine learning. The core of clustering resides in incorporating prior knowledge to construct supervision signals. According to different choices of supervision signals, one could roughly divide the evolution of clustering methods into three eras, \textit{i.e.}, classic clustering, deep clustering, and self-supervised clustering as depicted in Fig~\ref{fig: performance}. At the early stage, classic clustering methods build upon various assumptions on the data distribution, such as compactness~\cite{Kmeans, DBSCAN}, hierarchy~\cite{AgglomerativeClustering}, connectivity~\cite{SpectralClustering, SpectralClusteringNie, wang2020large}, sparsity~\cite{SSC, liu2017sparse}, and low rank~\cite{NMF, LRR, nie2016constrained}. Though having achieved promising performance, classic clustering methods would produce suboptimal results confronting complex and high-dimensional data. As an improvement, deep clustering methods equip clustering models with neural networks to extract discriminative features~\cite{DeepClusteringPeng2016, JULE, DEC, deepclustering_wangqi}. In alignment with priors such as cluster discriminability~\cite{DeepClusteringGhasedi} and balance~\cite{IMSAT}, various supervision signals are formulated to optimize the clustering network. In the last few years, motivated by the success of self-supervised learning~\cite{MOCO, SimCLR, BYOL}, clustering methods turn to creating supervision signals through data augmentation~\cite{CC, SCAN, dang2021nearest} or momentum strategies~\cite{GCC, ProPos}.

Though varying in the method design, most existing clustering methods design supervision signals in an internal manner. Despite the remarkable success achieved, the internally guided clustering paradigm faces an inherent limitation. Specifically, the hand-crafted internal supervision signals, even enhanced with data augmentation, are inherently upper-bounded by the limited information in the given data. For example, ``Corgi'' and ``Shiba Inu'' dogs are visually similar and are likely to be confused in image clustering. Luckily, beyond the internal signals, we notice there also exists well-established external knowledge that potentially conduces to clustering, while having been regrettably and largely ignored. In the above example, we could better distinguish the images given the external knowledge that ``Corgi'' have shorter and thicker legs compared with ``Shiba Inu'' dogs. In short, from different sources or modalities, the external knowledge could serve as promising supervision signals to guide clustering. Compared with exhaustively mining internal supervision signals from data, it would yield twice the effect with half the effort by incorporating rich and readily available external knowledge to guide clustering.

\begin{figure}[t]
\centering
\includegraphics[width=0.85\columnwidth]{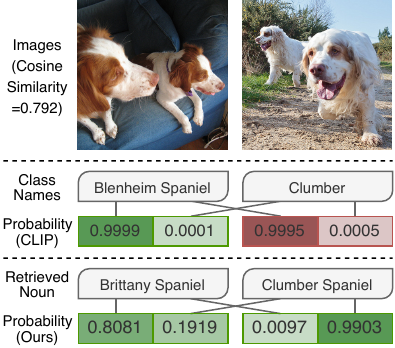}
\caption{Our observations with two image examples from the ImageNet-Dogs dataset as a showcase. For each example, we show the manually annotated class names and the nouns obtained by the proposed TAC, as well as the zero-shot classification probabilities. From the example, one could arrive at two observations, namely, i) visually similar samples could be better distinguished in the text modality, and ii) manually annotated class names are not always the best semantic description. As shown, zero-shot CLIP falsely classifies both images to the \textit{Blenheim Spaniel} class (probably due to the word \textit{Spaniel}), whereas the nouns obtained by our TAC successfully separate them. Such observations suggest a great opportunity to leverage the external knowledge (hidden in the text modality in this showcase) to facilitate image clustering.} \label{fig: example}
\end{figure}

In this work, we propose a simple yet effective externally guided clustering method TAC (Text-Aided Clustering), which clusters images by incorporating external knowledge from the text modality. In the absence of class name priors, there are two challenges in leveraging the textual semantics for image clustering, namely, \textit{i)} how to construct the text space, and \textit{ii)} how to collaborate images and texts for clustering. For the first challenge, ideally, we expect the text counterparts of between-class images to be highly distinguishable so that clustering can be easily achieved. To this end, inspired by the zero-shot classification paradigm in CLIP~\cite{CLIP}, we reversely classify all nouns from WordNet~\cite{WordNet} to image semantic centers. Based on the classification confidence, we select the most discriminative nouns for each image center to form the text space and retrieve a text counterpart for each image. Intriguingly, Fig.~\ref{fig: example} demonstrates that in certain cases, the retrieved nouns could describe the image semantics, sometimes even better than the manually annotated class names. For the second challenge, we first establish an extremely simple baseline by concatenating the images and text counterparts, which already significantly enhances the k-means clustering performance without any additional training. For better collaboration, we propose to mutually distill the neighborhood information between the text and image modalities. By additionally training cluster heads, the proposed TAC achieves state-of-the-art performance on five widely used and three more challenging image clustering datasets. Without loss of generality, we evaluate TAC on the pre-trained CLIP model in our experiments, but TAC could adapt to any vision-language pre-trained (VLP) model by design.

The major contributions of this work could be summarized as follows:
\begin{itemize}
	\item Unlike previous clustering works that exhaustively explore and exploit supervision signals internally, we propose leveraging external knowledge to facilitate clustering. We summarize such a novel paradigm as externally guided clustering, which provides an innovative perspective on the construction of supervision signals.
	\item To implement and validate our idea, we propose an externally guided clustering method TAC, which leverages the textual semantics to facilitate image clustering. Experiments demonstrate the superiority of TAC over eight datasets, including ImageNet-1K. Impressively, in most cases, TAC even outperforms zero-shot CLIP in the absence of class name priors.
	\item The significance of TAC is two-fold. On the one hand, it proves the effectiveness and superiority of the proposed externally guided clustering paradigm. On the other hand, it suggests the presence of more simple but effective strategies for mining the zero-shot learning ability inherent in VLP models.
\end{itemize}

\section{Related Work}
In this section, we review the deep clustering methods, and the zero-shot classification paradigm of VLP models which also utilizes the text modality to perform visual tasks.

\subsection{Deep Image Clustering}
In addition to effective clustering strategies, discriminative features also play an important role in clustering. Benefiting from the powerful feature extraction ability of neural networks, deep clustering methods show their superiority in handling complex and high-dimensional data~\cite{DeepClusteringPeng2016, DeepClusteringGuo, DeepClusteringGhasedi}. The pioneers in deep clustering focus on learning clustering-favorable features through optimizing the network with clustering objectives~\cite{JULE, DEC, DeepClusteringPeng2018, IMSAT, PICA, IIC}. In recent years, motivated by the success of contrastive learning, a series of contrastive clustering methods achieve substantial performance leaps on image clustering benchmarks~\cite{CC, TCC, GCC}. Instead of clustering images in an end-to-end manner, several works initially learn image embedding through uni-modal pre-training and subsequently mine clusters based on neighborhood consistency~\cite{SCAN, NNM} or pseudo-labeling~\cite{SPICE}. By disentangling representation learning and clustering, these multi-stage methods enjoy higher flexibility for their easy adaption to superior pre-trained models. A recent study~\cite{ProPos} demonstrates that the clustering performance could be further improved when equipping clustering models with more advanced representation learning methods~\cite{BYOL}. Very recently, SIC~\cite{SIC} attempts to generate image pseudo labels from the textual space.

Though having achieved remarkable progressions, almost all existing deep image clustering methods mine supervision signals internally. However, the internal supervision signals are inherently bounded by the given images. Instead of pursuing internal supervision signals following previous works, we propose a new paradigm that leverages external knowledge to facilitate image clustering. We hope the simple design and engaging performance of TAC could attract more attention to the externally guided clustering paradigm.

\subsection{Zero-shot Classification}
Recently, more and more efforts have been devoted to multi-modal, especially vision-language pre-training (VLP). By learning the abundant image-text pairs on the Internet, VLP methods~\cite{UNIMO, Beit3}, have achieved impressive performance in multi-modal representation learning. More importantly, unlike uni-modal pre-trained models that require additional fine-tuning, VLP models could adapt to various tasks such as classification~\cite{CLIP}, segmentation~\cite{DenseCLIP}, and image captioning~\cite{BLIP} in a zero-shot manner. Here, we briefly introduce the zero-shot image classification paradigm in CLIP~\cite{CLIP} as an example. Given names of $K$ classes, CLIP first assembles them with prompts like ``a photo of [CLASS]'', where the [CLASS] token is replaced by the specific class name. Then, CLIP computes the text embeddings $\{w_i\}_{i=1}^K$ of the prompted sentences with its pre-trained text encoder. Finally, CLIP treats the embeddings $\{w_i\}_{i=1}^K$ as the classifier weight, and predicts the probability of image $\textbf{v}$ belonging to the $i$-th class as
\begin{equation}
\label{eq: probability}
	p(y=i | \textbf{v}) = \frac{\exp(\operatorname{sim}(v, w_i)/\tau)}{\sum_{j=1}^K \exp(\operatorname{sim}(v, w_j)/\tau)},
\end{equation}
where $v$ denotes the image features, $\operatorname{sim}(\cdot, \cdot)$ refers to the cosine similarity, and $\tau$ is the learned \textit{softmax} temperature.

Thanks to the consistent form between pre-training and inference, CLIP achieves promising results in zero-shot image classification. However, such a paradigm requires prior knowledge of class names, which is unavailable in clustering. To leverage CLIP for image clustering, the most direct approach is performing k-means~\cite{Kmeans} on the image embeddings. Nevertheless, the performance of k-means is limited and the textual semantics are underutilized. In this work, we explore a more advanced paradigm for image clustering, by taking full advantage of both the pre-trained image and text encoders. Intriguingly, experiments demonstrate that even in the absence of class name priors, the proposed TAC outperforms zero-shot CLIP in most cases. We hope this work could bring some insights into the paradigm design of leveraging VLP models for downstream classification and clustering.

\begin{figure*}[t]
\centering
\includegraphics[width=\textwidth]{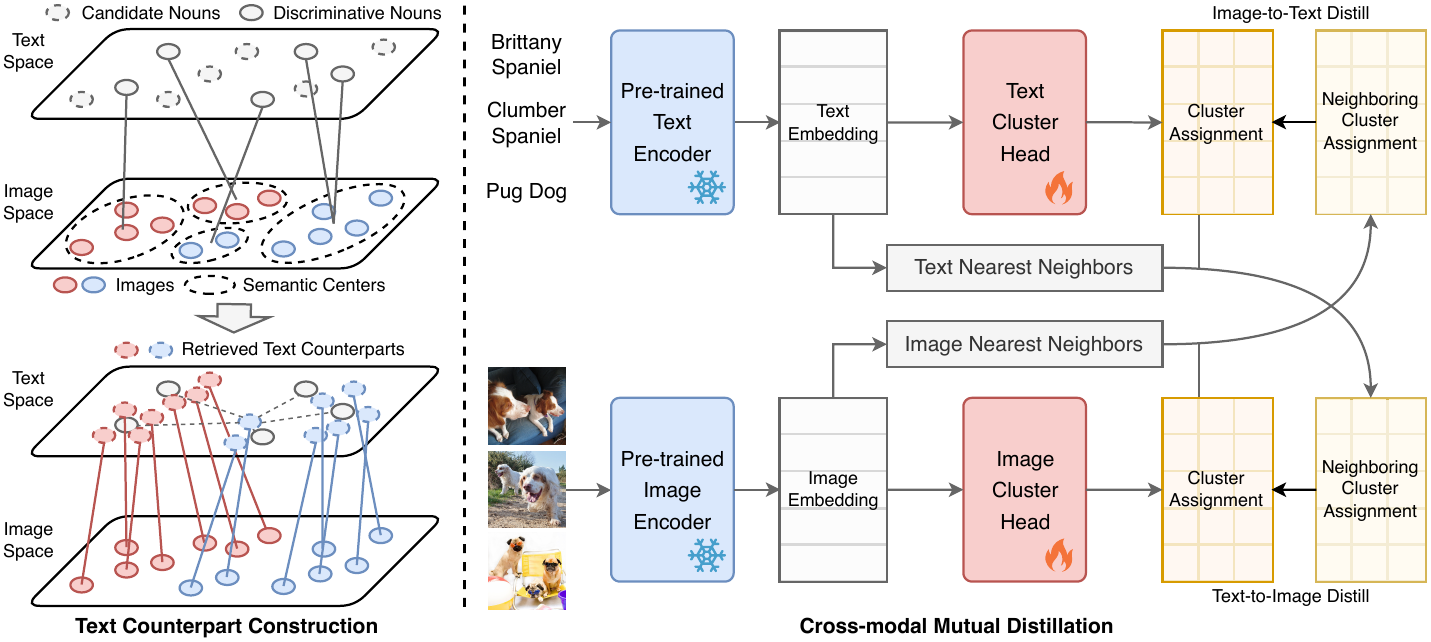}
\caption{Overview of the proposed TAC. \textbf{(Left)} TAC first classifies all nouns from WordNet to image semantic centers, and selects the most discriminative nouns to construct the text space. After that, TAC retrieves nouns for each image to compute its counterpart in the text space. By concatenating the image and retrieved text, we arrive at an extremely simple baseline without any additional training. \textbf{(Right)} To better collaborate the text and image modalities, TAC trains cluster heads by mutually distilling the neighborhood information. In brief, TAC encourages images to have consistent cluster assignments with the nearest neighbors of their counterparts in the text embedding space, and vice versa. Such a cross-modal mutual distillation strategy further boosts the clustering performance of TAC.}\label{fig: method}
\end{figure*}

\section{Method}

In this section, we present TAC, a simple yet effective externally guided clustering method illustrated in Fig.~\ref{fig: method}. In brief, we first propose a text counterpart construction strategy to exploit the text modality in Sec.~\ref{sec: semantic}. Then, we propose a cross-modal mutual distillation strategy to collaborate the text and image modalities in Sec.~\ref{sec: distill}.

\subsection{Text Counterpart Construction}
\label{sec: semantic}
The textual semantics are naturally favored in discriminative tasks such as classification and clustering. Ideally, clustering could be easily achieved if images have highly distinguishable counterparts in the text modality. To this end, in the absence of class name priors, we propose to select a subset of nouns from WordNet~\cite{WordNet} to compose the text space, which is expected to exhibit the following two merits, namely, \textit{i}) precisely covering the image semantics; and \textit{ii}) highly distinguishable between images of different semantics.

The image semantics of different granularities could be captured by k-means with various choices of $k$. A small value of $k$ corresponds to coarse-grained semantics, which might not be precise enough to cover the semantics of images at cluster boundaries. Oppositely, a large value of $k$ produces fine-grained semantics, which might fail to distinguish images from different classes. To find image semantics of appropriate granularity, we estimate $k=N / 300$ given $N$ images, hypothesizing a cluster of $\tilde{N}=300$ images is compact enough to be described by the same set of nouns. Experiments in Section~\ref{sec: cluster size} show that our TAC is robust across a reasonable range of $\tilde{N}$. With the estimated value of $k$, we apply k-means on image embeddings to compute the image semantic centers by
\begin{equation}
\label{eq: image center}
	s_l = \sum_{i=1}^N \mathbbm{1}_{v_i \in l}~v_i, l \in [1, k],
\end{equation}
where $\mathbbm{1}_{v_i \in l}$ is the indicator which equals one iff image $v_i$ belongs to the $l$-th cluster.

Next, we aim to find discriminative nouns to describe each semantic center. Here, motivated by the zero-shot classification paradigm of CLIP, we reversely classify all nouns from WordNet into $k$ image semantic centers. Specifically, the probability of the $i$-th noun belonging to the $l$-th image semantic center is
\begin{equation}
\label{eq: noun classification}
	p(y=l | \mathbf{t_i}) = \frac{\exp(\operatorname{sim}(t_i, s_l))}{\sum_{j=1}^k \exp(\operatorname{sim}(t_i, s_j))},
\end{equation}
where $\mathbf{t_i}$ denoted the $i$-th noun prompted like CLIP, and $t_i$ is the feature extracted by the text encoder. To identify highly representative and distinguishable nouns, we select the top $\gamma$ confident nouns for each image semantic center. Formally, the $i$-th noun would be select for the $l$-th center if
\begin{gather}
	p(y=k | \mathbf{t_i}) \geq \bar{p}(y=k), \label{eq: noun selection} \\ 
	\bar{p}(y=k) = \operatorname{sort}\{p(y=k | \mathbf{t_i}) | \operatorname{argmax}p(y | \mathbf{t_i})=k \}[\gamma], \notag
\end{gather}
where $\bar{p}(y=k)$ corresponds to the $\gamma$-th largest confidence of nouns belonging to the $l$-th center. In practice, we fix $\gamma=5$ on all datasets.

The selected nouns compose the text space catering to the input images. Then, we retrieve nouns for each image to compute its counterpart in the text modality. To be specific, let $\{\mathbf{\bar{t}_i}\}_{i=1}^{M}$ be the set of $M$ selected nouns with $\{\bar{t}_i\}_{i=1}^{M}$ being their text embeddings, we compute the text counterpart $\tilde{t}_i$ for image $v_i$ as
\begin{gather}
	\tilde{t}_i = \sum_{j=1}^{M} p(\bar{t}_j | v_i) \bar{t}_j, \label{eq: noun retrieval 1} \\
	p(\bar{t}_j | v_i) = \frac{\exp(\operatorname{sim}(v_i, \bar{t}_j) / \tilde{\tau})}{\sum_{k=1}^M \exp(\operatorname{sim}(v_i, \bar{t}_k) / \tilde{\tau})}, \label{eq: noun retrieval 2}
\end{gather}
where $\tilde{\tau}=0.005$ controls the softness of retrieval. The design of soft retrieval is to prevent the text counterparts of different images from collapsing to the same point. After the text counterpart construction, we arrive at an extremely simple baseline by applying k-means on the concatenated features $[\tilde{t}_i, v_i]_{i=1}^N$. Notably, such an implementation requires no additional training or modifications on CLIP, but it could significantly improve the clustering performance compared with directly applying k-means on the image embeddings (see Section~\ref{sec: main results}).

\subsection{Cross-modal Mutual Distillation}
\label{sec: distill}
Though concatenating text counterparts and image embeddings improves the k-means performance, it is suboptimal for collaborating the two modalities. To better utilize multi-modal features, we propose the cross-modal mutual distillation strategy. Specifically, let $\mathcal{N}(v_i)$ be a random nearest neighbor of $v_i$, we introduce a cluster head $f: v \rightarrow p \in \mathcal{R}^K$ to predict the soft cluster assignments for images $v_i$ and $\mathcal{N}(v_i)$, where $K$ is the target cluster number. Formally, we denote the soft cluster assignments for $n$ images and their neighbors as
\begin{equation}
\label{eq: image assignments}
	P=\left[
		\begin{array}{c}
		p_1 \\
		\cdots \\
		p_n
		\end{array}
	\right] \text { and } 
	P^\mathcal{N}=\left[
		\begin{array}{c}
		p^\mathcal{N}_1 \\
		\cdots \\
		p^\mathcal{N}_n
		\end{array}
	\right].
\end{equation}
Likewise, we introduce another cluster head $g: \tilde{t}_i \rightarrow q_i \in \mathcal{R}^K$ to predict the soft cluster assignments for text counterpart $\tilde{t}_i$ and its random nearest neighbor $\mathcal{N}(\tilde{t}_i)$, resulting in the cluster assignment matrices
\begin{equation}
\label{eq: text assignments}
	Q=\left[
		\begin{array}{c}
		q_1 \\
		\cdots \\
		q_n
		\end{array}
	\right] \text { and } 
	Q^\mathcal{N}=\left[
		\begin{array}{c}
		q^\mathcal{N}_1 \\
		\cdots \\
		q^\mathcal{N}_n
		\end{array}
	\right].
\end{equation}

Let $\hat{p}_i, \hat{p}^\mathcal{N}_i, \hat{q}_i, \hat{q}^\mathcal{N}_i$ be the $i$-th column of assignment matrices $P, P^\mathcal{N}, Q, Q^\mathcal{N}$, the cross-modal mutual distillation loss is defined as follows, namely,
\begin{gather}
	L_{\mathrm{Dis}} = \sum_{i=1}^K L^{v\rightarrow t} _i + L^{t\rightarrow v}_i, \label{eq: distill} \\
	L^{v\rightarrow t} _i = -\log \frac{e^{(\operatorname{sim}(\hat{q}_i, \hat{p}^\mathcal{N}_i) / \hat{\tau})}}{\underset{k}{\sum} e^{(\operatorname{sim}(\hat{q}_i, \hat{p}^\mathcal{N}_k) / \hat{\tau})} + \underset{k\neq i}{\sum} e^{(\operatorname{sim}(\hat{q}_i, \hat{q}_k) / \hat{\tau})}  }, \label{eq: distill t} \\
	L^{t\rightarrow v} _i = -\log \frac{e^{(\operatorname{sim}(\hat{p}_i, \hat{q}^\mathcal{N}_i) / \hat{\tau})}}{\underset{k}{\sum} e^{(\operatorname{sim}(\hat{p}_i, \hat{q}^\mathcal{N}_k) / \hat{\tau})} + \underset{k\neq i}{\sum} e^{(\operatorname{sim}(\hat{p}_i, \hat{p}_k) / \hat{\tau})} }, \label{eq: distill v}
\end{gather}
where $\hat{\tau}$ is the \textit{softmax} temperature parameter. The distillation loss $L_{\mathrm{Dis}}$ has two effects. On the one hand, it minimizes the between-cluster similarity, leading to more discriminative clusters. On the other hand, it encourages consistent clustering assignments between each image and the neighbors of its text counterpart, and vice versa. In other words, it mutually distills the neighborhood information between the text and image modalities, bootstrapping the clustering performance in both. In practice, we set the number of nearest neighbors $\hat{N}=50$ on all datasets. Note that the neighbors are only computed once on all samples before training.

Next, we introduce two regularization terms to stabilize the training. First, to encourage the model to produce more confident cluster assignments, we introduce the following confidence loss, namely,
\begin{equation}
\label{eq: confidence}
	L_{\mathrm{Con}} = - \log \sum_{i=1}^n p_i^\top q_i,
\end{equation}
which would be minimized when both $p_i$ and $q_i$ become one-hot. Second, to prevent all samples from collapsing into only a few clusters, we adopt the balance loss, \textit{i.e.},
\begin{gather}
L_{\mathrm{Bal}} = - \sum_{i=1}^K \left(\bar{p}_i \log \bar{p}_i + \bar{q}_i \log \bar{q}_i\right), \label{eq: balance} \\
	\bar{p} = \frac{1}{n} \sum_{i=1}^n p_i \in \mathcal{R}^K, \bar{q} = \frac{1}{n} \sum_{i=1}^n q_i \in \mathcal{R}^K, \label{eq: balance pq}
\end{gather}
where $\bar{p}$ and $\bar{q}$ correspond to the cluster assignment distribution in the image and text modality, respectively.

Finally, we arrive at the overall objective function of TAC, which lies in the form of
\begin{equation}
\label{eq: overall loss}
	L_{\mathrm{TAC}} = L_{\mathrm{Dis}} + L_{\mathrm{Con}} - \alpha \cdot L_{\mathrm{Bal}},
\end{equation}
where $\alpha=5$ is the weight parameter.

\section{Experiments}
In this section, we evaluate the proposed TAC on five widely used and three more challenging image clustering datasets. A series of quantitative and qualitative comparisons, ablation studies, and hyper-parameter analyses are carried out to investigate the effectiveness and robustness of the method.

\subsection{Experimental Setup}
We first introduce the datasets and metrics used for evaluation, and then provide the implementation details of TAC. 

\subsubsection{Datasets}

To evaluate the performance of our TAC, we first apply it to five widely-used image clustering datasets including STL-10~\cite{STL}, CIFAR-10~\cite{CIFAR}, CIFAR-20~\cite{CIFAR}, ImageNet-10~\cite{ImageNet_10_Dogs}, and ImageNet-Dogs~\cite{ImageNet_10_Dogs}. With the rapid development of pre-training and clustering methods, we find clustering on relatively simple datasets such as STL-10 and CIFAR-10 is no longer challenging. Thus, we further evaluate the proposed TAC on three more complex datasets with larger cluster numbers, including DTD~\cite{DTD}, UCF-101~\cite{UCF101}, and ImageNet-1K~\cite{ImageNet}. Following recent deep clustering works~\cite{SCAN, NNM}, we train and evaluate TAC on the train and test splits, respectively. The brief information of all datasets used in our evaluation is summarized in Table~\ref{tab: dataset}.

\begin{table}[h]
\centering
\caption{A summary of datasets used for evaluation.}
\label{tab: dataset}
\resizebox{\columnwidth}{!}{%
\begin{tabular}{@{}lccccc@{}}
\toprule
Dataset       & Training Split & Test Split & \# Training & \# Test & \# Classes \\ \midrule
STL-10        & Train          & Test       & 5,000       & 8,000   & 10         \\
CIFAR-10      & Train          & Test       & 50,000      & 10,000  & 10         \\
CIFAR-20      & Train          & Test       & 50,000      & 10,000  & 20         \\
ImageNet-10   & Train          & Val        & 13,000      & 500     & 10         \\
ImageNet-Dogs & Train          & Val        & 19,500      & 750     & 15         \\ \midrule
DTD           & Train+Val      & Test       & 3,760       & 1,880   & 47   \\
UCF-101       & Train          & Val        & 9,537       & 3.783   & 101   \\
ImageNet-1K      & Train          & Val        & 1,281,167   & 50,000  & 1,000       \\ \bottomrule
\end{tabular}%
}
\end{table}

\begin{table*}[t]
\caption{Clustering performance on five widely-used image clustering datasets. The best and second best results are denoted in \textbf{bold} and \underline{underline}, respectively.}
\label{tab: classic}
\centering
\resizebox{\textwidth}{!}{%
\begin{tabular}{@{}lcccccccccccccccc@{}}
\toprule
Dataset & \multicolumn{3}{c}{STL-10} & \multicolumn{3}{c}{CIFAR-10} & \multicolumn{3}{c}{CIFAR-20} & \multicolumn{3}{c}{ImageNet-10} & \multicolumn{3}{c}{ImageNet-Dogs} & \multirow{2}{*}{AVG} \\ \cmidrule(r){1-16}
Metrics          & NMI  & ACC  & ARI  & NMI  & ACC  & ARI  & NMI  & ACC  & ARI  & NMI  & ACC  & ARI  & NMI  & ACC  & ARI  & \\ \midrule

JULE~\cite{JULE}             & 18.2 & 27.7 & 16.4 & 19.2 & 27.2 & 13.8 & 10.3 & 13.7 & 3.3  & 17.5 & 30.0 & 13.8 & 5.4  & 13.8 & 2.8  & 15.5 \\
DEC~\cite{DEC}              & 27.6 & 35.9 & 18.6 & 25.7 & 30.1 & 16.1 & 13.6 & 18.5 & 5.0  & 28.2 & 38.1 & 20.3 & 12.2  & 19.5 & 7.9  & 21.2 \\
DAC~\cite{DAC}              & 36.6 & 47.0 & 25.7 & 39.6 & 52.2 & 30.6 & 18.5 & 23.8 & 8.8  & 39.4 & 52.7 & 30.2 & 21.9 & 27.5 & 11.1 & 31.0 \\
DCCM~\cite{DCCM}             & 37.6 & 48.2 & 26.2 & 49.6 & 62.3 & 40.8 & 28.5 & 32.7 & 17.3 & 60.8 & 71.0 & 55.5 & 32.1 & 38.3 & 18.2 & 41.3 \\
IIC~\cite{IIC}              & 49.6 & 59.6 & 39.7 & 51.3 & 61.7 & 41.1 & 22.5 & 25.7 & 11.7 & --   & --   & --   & --   & --   & --   & -- \\
PICA~\cite{PICA}             & 61.1 & 71.3 & 53.1 & 59.1 & 69.6 & 51.2 & 31.0 & 33.7 & 17.1 & 80.2 & 87.0 & 76.1 & 35.2 & 35.3 & 20.1 & 52.1 \\
CC~\cite{CC}               & 76.4 & 85.0 & 72.6 & 70.5 & 79.0 & 63.7 & 43.1 & 42.9 & 26.6 & 85.9 & 89.3 & 82.2 & 44.5 & 42.9 & 27.4 & 62.1 \\
IDFD~\cite{IDFD}             & 64.3 & 75.6 & 57.5 & 71.1 & 81.5 & 66.3 & 42.6 & 42.5 & 26.4 & 89.8 & 95.4 & 90.1 & 54.6 & 59.1 & 41.3 & 63.9 \\
SCAN~\cite{SCAN}             & 69.8 & 80.9 & 64.6 & 79.7 & 88.3 & 77.2 & 48.6 & 50.7 & 33.3 & --   & --   & --   & 61.2 & 59.3 & 45.7 & -- \\
MiCE~\cite{MiCE}             & 63.5 & 75.2 & 57.5 & 73.7 & 83.5 & 69.8 & 43.6 & 44.0 & 28.0 & --   & --   & --   & 42.3 & 43.9 & 28.6 & -- \\
GCC~\cite{GCC}              & 68.4 & 78.8 & 63.1 & 76.4 & 85.6 & 72.8 & 47.2 & 47.2 & 30.5 & 84.2 & 90.1 & 82.2 & 49.0 & 52.6 & 36.2 & 64.3 \\
NNM~\cite{NNM}              & 66.3 & 76.8 & 59.6 & 73.7 & 83.7 & 69.4 & 48.0 & 45.9 & 30.2 & --   & --   & --   & 60.4 & 58.6 & 44.9 & -- \\
TCC~\cite{TCC}              & 73.2 & 81.4 & 68.9 & 79.0 & 90.6 & 73.3 & 47.9 & 49.1 & 31.2 & 84.8 & 89.7 & 82.5 & 55.4 & 59.5 & 41.7 & 67.2 \\ 
SPICE~\cite{SPICE}            & 81.7 & 90.8 & 81.2 & 73.4 & 83.8 & 70.5 & 44.8 & 46.8 & 29.4 & 82.8 & 92.1 & 83.6 & 57.2 & 64.6 & 47.9 & 68.7 \\ 
SIC~\cite{SIC}              & \underline{95.3} & \underline{98.1} & \underline{95.9} & \textbf{84.7} & \textbf{92.6} & \textbf{84.4} & 59.3 & \underline{58.3} & \underline{43.9} & 97.0 & 98.2 & 96.1 & 69.0 & 69.7 & 55.8 & \underline{79.9} \\
CLIP (k-means)   & 91.7 & 94.3 & 89.1 & 70.3 & 74.2 & 61.6 & 49.9 & 45.5 & 28.3 & 96.9 & 98.2 & 96.1 & 39.8 & 38.1 & 20.1 & 66.3 \\ \midrule
TAC (no train)  & 92.3 & 94.5 & 89.5 & 80.8 & 90.1 & 79.8 & \underline{60.7} & 55.8 & 42.7 & \underline{97.5} & \underline{98.6} & \underline{97.0} & \underline{75.1} & \underline{75.1} & \underline{63.6} & 79.5 \\
TAC             & \textbf{95.5} & \textbf{98.2} & \textbf{96.1} & \underline{83.3} & \underline{91.9} & \underline{83.1} & \textbf{61.1} & \textbf{60.7} & \textbf{44.8} & \textbf{98.5} & \textbf{99.2} & \textbf{98.3} & \textbf{80.6} & \textbf{83.0} & \textbf{72.2} & \textbf{83.2} \\
\rowstyle{\color{hr}} CLIP (zero-shot) & \rowstyle{\color{hr}}93.9 & \rowstyle{\color{hr}}97.1 & \rowstyle{\color{hr}}93.7 & \rowstyle{\color{hr}}80.7 & \rowstyle{\color{hr}}90.0 & \rowstyle{\color{hr}}79.3 & \rowstyle{\color{hr}}55.3 & \rowstyle{\color{hr}}58.3 & \rowstyle{\color{hr}}39.8 & \rowstyle{\color{hr}}95.8 & \rowstyle{\color{hr}}97.6 & \rowstyle{\color{hr}}94.9 & \rowstyle{\color{hr}}73.5 & \rowstyle{\color{hr}}72.8 & \rowstyle{\color{hr}}58.2 & \rowstyle{\color{hr}}78.7 \\
\bottomrule
\end{tabular}%
}
\end{table*}

\subsubsection{Evaluation metrics}
We adopt three widely-used metrics including Normalized Mutual Information (NMI), Accuracy (ACC), and Adjusted Rand Index (ARI) to evaluate the clustering performance. Higher values of these metrics indicate better results.

\subsubsection{Implementation details}

Following previous works~\cite{SIC}, we adopt the pre-trained CLIP model with ViT-B/32~\cite{ViT} and Transformer~\cite{Transformer} as image and text backbones, respectively. For nouns from WordNet~\cite{WordNet}, we assemble them with prompts like ``A photo of [CLASS]'' before feeding them into the Transformer. For datasets with an average cluster size less than 300, we empirically set $k$ in k-means thrice as the target cluster number $K$. The two cluster heads $f$ and $g$ are two-layer MLPs of dimension $512$-$512$-$K$. We train $f$ and $g$ by the Adam optimizer with an initial learning rate of $1e{-3}$ for $20$ epochs, with a batch size of $512$. We fix $\tau = 5e{-3}$, $\hat{\tau}=0.5$, and $\alpha=5.0$ in all the experiments. The only exception is that on UCF-101 and ImageNet-1K, we change $\hat{\tau}$ to $5.0$, batch size to $8192$, and training epochs to $100$, catering to the large cluster number. All experiments are conducted on a single Nvidia RTX 3090 GPU. In our experiments, it takes only one minute to train TAC on the CIFAR-10 dataset.

\subsection{Main Results}
\label{sec: main results}
Here we compare TAC with state-of-the-art baselines on five classic and three more challenging image clustering datasets, followed by feature visualizations to show the superiority of the proposed TAC.

\subsubsection{Performance on classic datasets}

We first evaluate the proposed TAC on five widely-used image clustering datasets, compared with 15 deep clustering baselines. While early baselines adopt ResNet-34(18) as the backbone, here we mainly focus on comparisons with zero-shot CLIP and CLIP-based methods. As shown in Table~\ref{tab: classic}, by simply retrieving a text counterpart for each image, the proposed TAC successfully mines ``free'' semantic information from the text encoder. Without any additional training, TAC (no train) substantially improves the k-means clustering performance, especially on more complex datasets. For example, it achieves 14.4\% and 43.5\% ARI improvements on CIFAR-20 and ImageNet-Dogs, respectively. When further enhanced with the proposed cross-modal mutual distillation strategy, TAC achieves state-of-the-art clustering performance, even surpassing zero-shot CLIP on all five datasets. Such compelling results demonstrate that beyond the current zero-shot classification paradigm, alternative simple but more effective strategies exist for mining the VLP model's ability in image classification and clustering.

\begin{table*}[t]
\centering
\caption{Clustering performance on three more challenging image clustering datasets. The best and second best results are denoted \\ in \textbf{bold} and \underline{underline}, respectively.}
\label{tab: complex}
\resizebox{0.8\textwidth}{!}{%
\begin{tabular}{@{}*l^c^c^c^c^c^c^c^c^c^c@{}}
\toprule
Dataset & \multicolumn{3}{c}{DTD} & \multicolumn{3}{c}{UCF-101} & \multicolumn{3}{c}{ImageNet-1K} & \multirow{2}{*}{AVG} \\ \cmidrule(r){1-10}
Metrics          & NMI  & ACC  & ARI  & NMI  & ACC  & ARI  & NMI  & ACC  & ARI  & \\ \midrule
CLIP (k-means)~\cite{CLIP}   & 57.3 & 42.6 & 27.4 & 79.5 & 58.2 & 47.6 & 72.3 & 38.9 & 27.1 & 50.1 \\
SCAN~\cite{SCAN}            & 59.4 & \underline{46.4} & \underline{31.7} & 79.7 & 61.1 & 53.1 & 74.7 & 44.7 & 32.4 & 53.7 \\
SIC~\cite{SIC}              & 59.6 & 45.9 & 30.5 & 81.0 & \underline{61.9} & \underline{53.6} & 77.2 & 47.0 & 34.3 & 54.6 \\ \midrule
TAC (no train)  & \underline{60.1} & 45.9 & 29.0 & \underline{81.6} & 61.3 & 52.4 & \underline{77.8} & \underline{48.9} & \underline{36.4} & \underline{54.8} \\
TAC             & \textbf{62.1} & \textbf{50.1} & \textbf{34.4} & \textbf{82.3} & \textbf{68.7} &      \textbf{60.1} & \textbf{79.9} & \textbf{58.2} & \textbf{43.5} & \textbf{59.9} \\
\rowstyle{\color{hr}} CLIP (zero-shot)~\cite{CLIP} & 56.5 & 43.1 & 26.9 & 79.9 & 63.4 & 50.2 & 81.0 & 63.6 & 45.4 & 56.7 \\ \bottomrule
\end{tabular}
}
\end{table*}

\begin{figure*}[t]
\centering
  \subfigure[CLIP Image Embedding]{
    \includegraphics[width=0.235\textwidth]{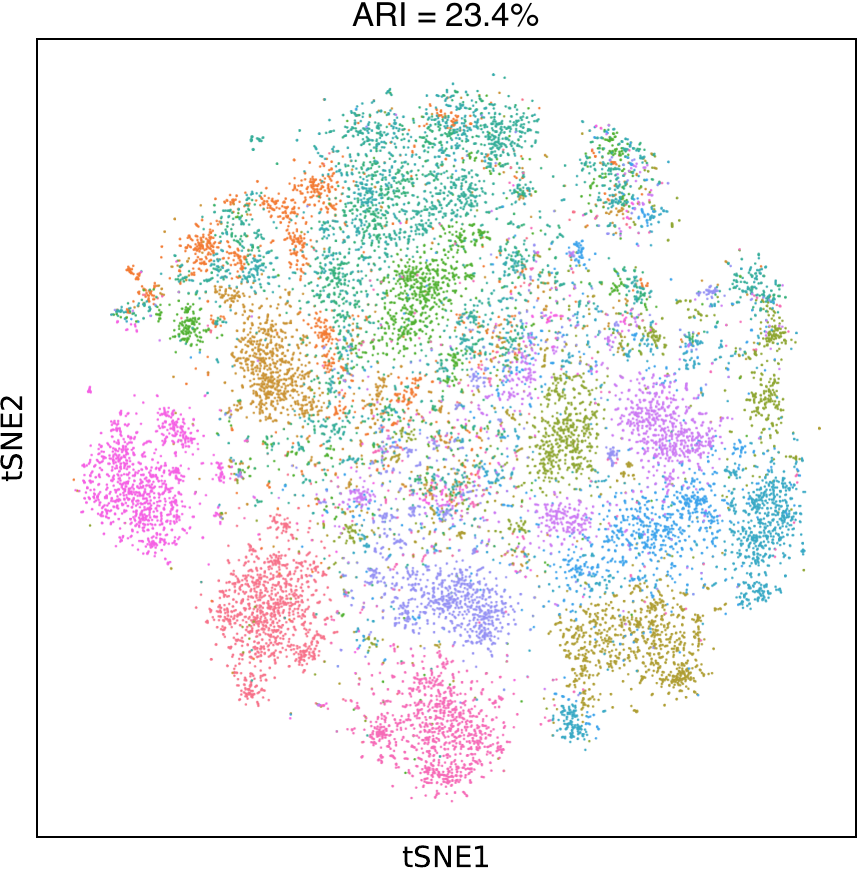}\label{fig: clip image}}
  \subfigure[Constructed Text Counterpart]{
    \includegraphics[width=0.235\textwidth]{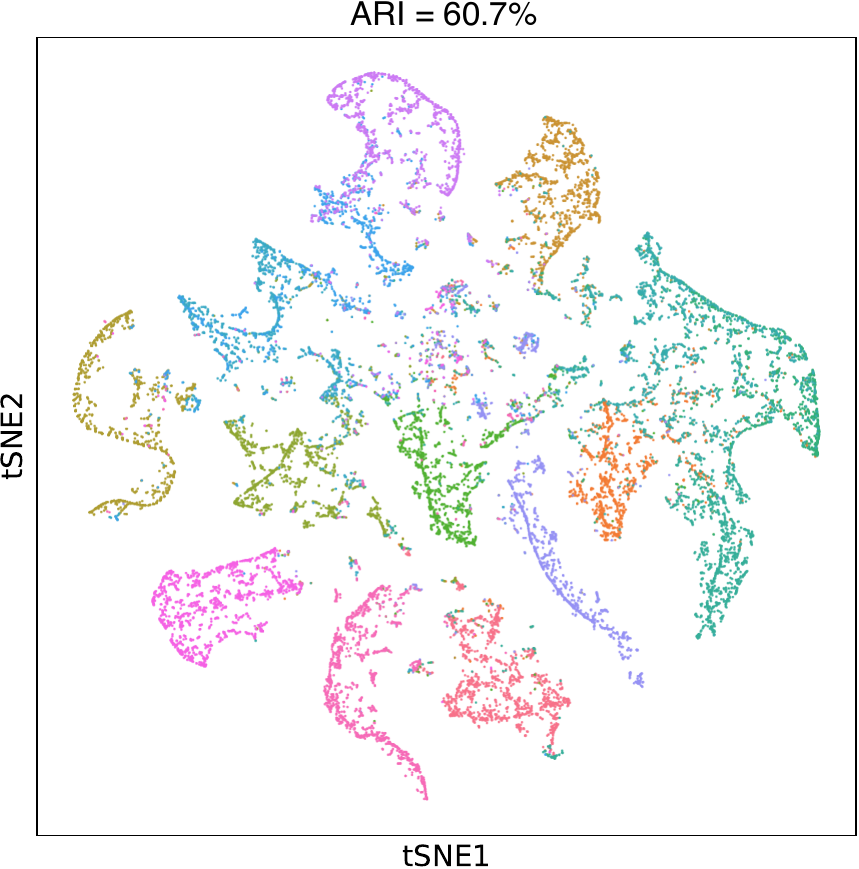}\label{fig: retrieved text}}
  \subfigure[TAC (no train)]{
    \includegraphics[width=0.235\textwidth]{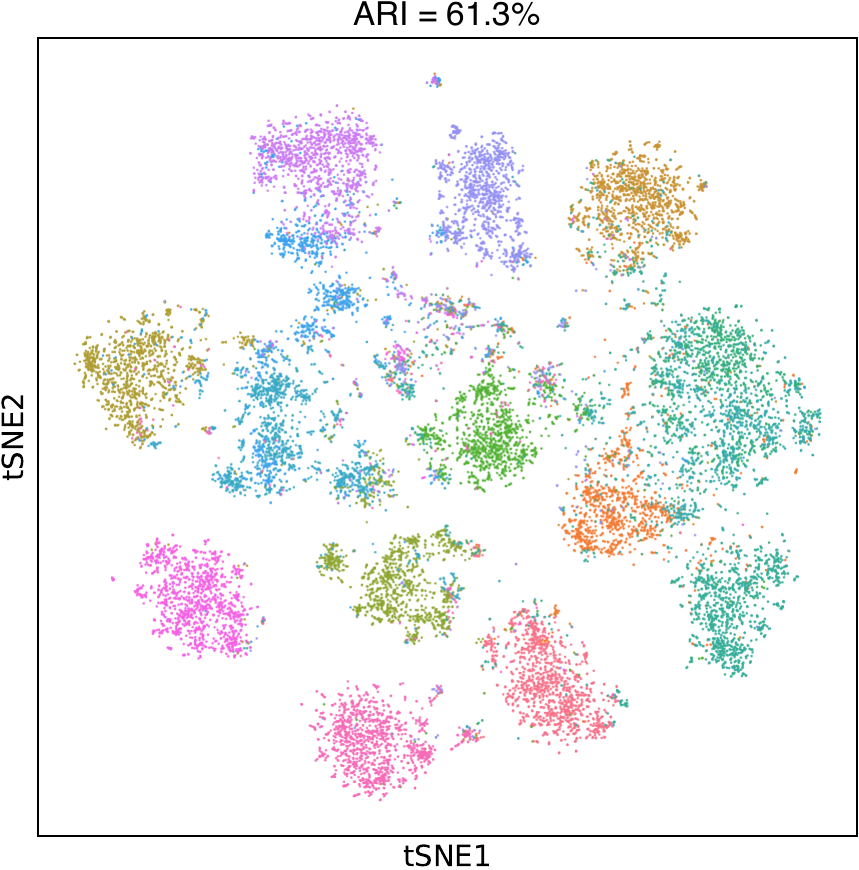}\label{fig: concat}}
  \subfigure[TAC]{
    \includegraphics[width=0.235\textwidth]{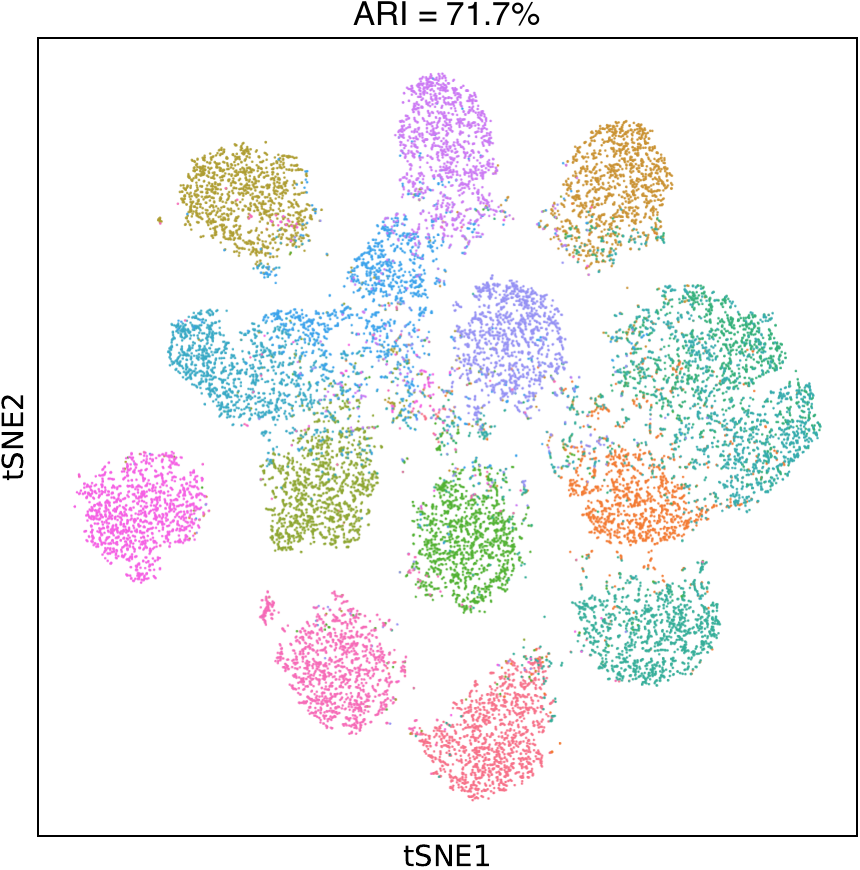}\label{fig: TAC}}
  \caption{Visualization of features extracted by different methods on the ImageNet-Dogs training set, with the corresponding k-means clustering ARI annotated on the top. a) image embedding directly obtained from the CLIP image encoder; b) text counterparts constructed by TAC; c) concatenation of images and text counterparts; d) representation learned by TAC through cross-modal mutual distillation.}\label{fig: tsne}
\end{figure*}

\subsubsection{Performance on challenging datasets}

The clustering results of TAC and baseline methods on more challenging datasets are provided in Table~\ref{tab: complex}. Firstly, we observe TAC without additional training could consistently boost the k-means performance, which achieves a 10\% improvement in clustering accuracy on ImageNet-1K. Secondly, although zero-shot CLIP yields slightly better performance on ImageNet-1K given its substantial prior knowledge of 1K class names, TAC still achieves superior performance on DTD and UCF-101 without the class name prior. Such a result verifies the effectiveness of the proposed text counterpart construction strategy, as well as our observation that manually annotated class names are not always the best semantic description.

\subsubsection{Visualization}

To provide an intuitive understanding of the clustering results, we visualize the features obtained at four different steps of TAC in Fig.~\ref{fig: tsne}. The clustering performance by applying k-means on the features is annotated at the top. Fig.~\ref{fig: clip image} shows the image features extracted by the pre-trained CLIP image encoder. As can be seen, images of different breeds of dogs are mixed, leading to a poor clustering ARI of 23.4\%. By selecting and retrieving discriminative nouns, visually similar samples could be better distinguished in the text modality as shown in Fig.~\ref{fig: retrieved text}. By simply concatenating images and retrieved text counterparts, TAC significantly improves the feature discriminability and k-means performance without any additional training. Finally, when equipped with the proposed cross-modal mutual distillation strategy, TAC could better collaborate the image and text modalities, leading to the best within-clustering compactness and between-cluster scatterness.

\subsection{Ablation Study}
In this section, we conduct ablation studies on the three loss terms and the direction of the cross-modal distillation.

\subsubsection{Loss terms}
To understand the efficacy of the three loss terms $L_{\mathrm{Dis}}$, $L_{\mathrm{Con}}$, and $L_{\mathrm{Bal}}$ in Eq.~\ref{eq: distill}, \ref{eq: confidence}, and \ref{eq: balance}, we evaluate the performance of TAC with different loss combinations. According to the results in Table~\ref{tab: loss}, one could see that: \textit{i)} the balance loss $L_{\mathrm{Bal}}$ could prevent cluster collapsing. Without $L_{\mathrm{Bal}}$, TAC assigns most images to only a few clusters, leading to poor clustering performance on both datasets; \textit{ii)} the confidence loss $L_{\mathrm{Con}}$ is necessary for datasets with large cluster numbers. The reason is that the cluster assignments would be less confident when the cluster number is large. In this case, the regularization efficacy of $L_{\mathrm{Bal}}$ would be alleviated, which explains the performance degradation on UCF-101; and \textit{iii)} $L_{\mathrm{Dis}}$ could effectively distill the neighborhood information between the text and image modalities, leading to the best clustering performance.
\begin{table}[h]
\centering
\caption{The performance of TAC with different combinations of the loss terms.}
\label{tab: loss}
\resizebox{0.95\columnwidth}{!}{%
\begin{tabular}{@{}ccccccccc@{}}
\toprule
\multirow{2}{*}{$L_{\mathrm{Dis}}$} & \multirow{2}{*}{$L_{\mathrm{Con}}$} & \multirow{2}{*}{$L_{\mathrm{Bal}}$} & \multicolumn{3}{c}{ImageNet-Dogs} & \multicolumn{3}{c}{UCF-101} \\ \cmidrule(l){4-9} 
  &   &   & NMI & ACC & ARI & NMI & ACC & ARI \\ \midrule
\ding{51} &   &   & 71.4 & 69.5 & 38.1 & 69.3 & 7.6 & 13.6 \\
  & \ding{51} &   & 57.2 &14.3 & 24.3 & 52.1 & 3.4 & 8.6 \\
  &   & \ding{51} & 15.1 & 19.3 & 4.1 & 43.5 & 16.2 &  5.7 \\
\ding{51} & \ding{51} &   & 72.5 & 57.0 & 45.3 & 55.6 & 3.6 & 9.9 \\
\ding{51} &   & \ding{51} & \textbf{80.6} & \textbf{83.5} & \textbf{72.3} & 70.5 & 45.1 & 34.5 \\
  & \ding{51} & \ding{51} & 78.2 & 81.8 & 69.6 & \underline{81.6} & \underline{67.3} & \underline{59.1} \\
\ding{51} & \ding{51} & \ding{51} & \textbf{80.6} & \underline{83.0} & \underline{72.2} & \textbf{82.3} & \textbf{68.7} & \textbf{60.1} \\ \bottomrule
\end{tabular}%
}
\end{table}

\subsubsection{Distillation direction}

Recall that the cross-modal distillation strategy mutually distills the neighborhood information from one modality to another. To better understand the effectiveness of mutual distillation, we evaluate the performance of TAC with different directions of the distillation in Table~\ref{tab: distill}. As can be seen, text-to-image distillation gives inferior performance compared with bi-directional distillation, probably due to the less exploration of image neighborhood information. Moreover, in the one-directional scenarios, text-to-image outperforms image-to-text distillation, which proves that the textual semantics are more favorable for clustering.

\begin{table}[h]
\centering
\caption{The performance of TAC with different distillation directions. $\dag$: Use the text head for clustering.}
\label{tab: distill}
\resizebox{0.95\columnwidth}{!}{%
\begin{tabular}{@{}ccccccc@{}}
\toprule
\multirow{2}{*}{Direction}     & \multicolumn{3}{c}{ImageNet-Dogs} & \multicolumn{3}{c}{UCF-101} \\ \cmidrule(l){2-7} 
                               & NMI       & ACC       & ARI       & NMI     & ACC     & ARI     \\ \midrule
Image $\rightarrow$ Text$^\dag$ & 76.5 & 79.3 & 67.1 & 78.5 & 64.1 & 53.7 \\
Text $\rightarrow$ Image       & 78.8 & 82.1 & 69.4 & 81.1 & 65.8 & 57.5 \\
Image $\leftrightarrow$ Text   & \textbf{80.6} & \textbf{83.0} & \textbf{72.2} & \textbf{82.3} & \textbf{68.7} & \textbf{60.1} \\ \bottomrule
\end{tabular}%
}
\end{table}

\subsection{Parameter Analyses}

\begin{figure*}[t]
\centering
  \subfigure[~~~Expected compact cluster size $\tilde{N}$]{
    \includegraphics[width=0.31\textwidth]{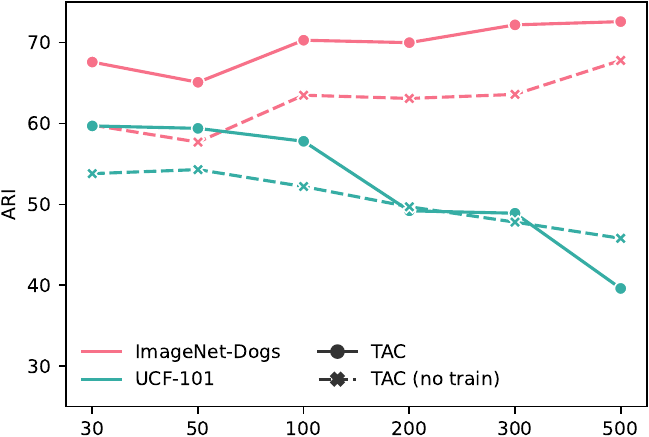}\label{fig: compact_cluster_size}}
  \subfigure[~~~Number of discriminative nouns $\gamma$]{
    \includegraphics[width=0.31\textwidth]{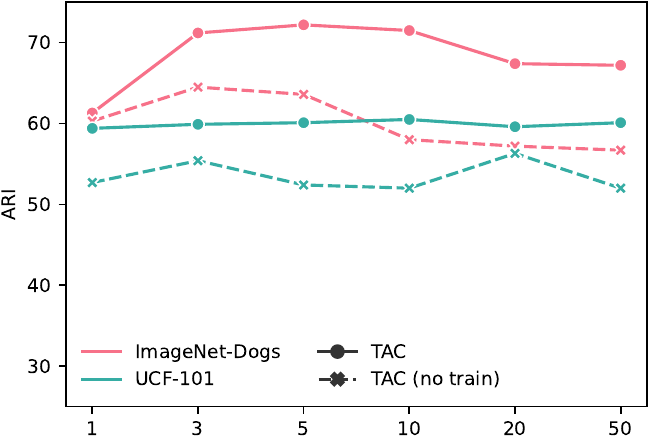}\label{fig: number_of_discriminative_nouns}}
  \subfigure[~~~Number of nearest neighbors $\hat{N}$]{
    \includegraphics[width=0.31\textwidth]{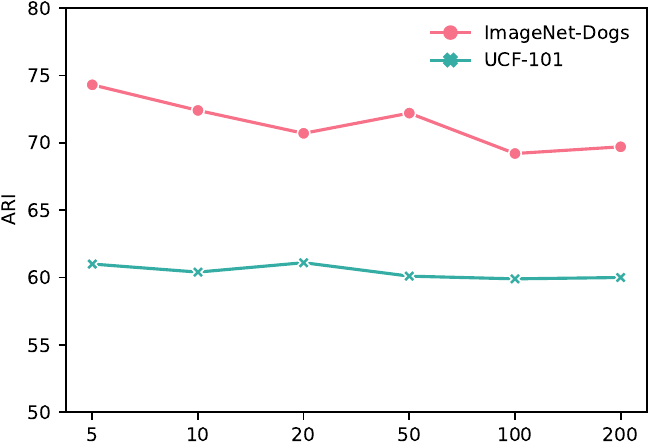}\label{fig: number_of_neighbors}}
  \caption{Analyses on three hyper-parameters in the proposed TAC. The first two hyper-parameters influence both TAC with and without training. The last hyper-parameter only influences the cross-modal mutual distillation process of TAC.}\label{fig: parameter}
\end{figure*}

To evaluate the robustness of TAC, we evaluate it under various choices of the expected compact cluster size $\tilde{N}$, the number of discriminative nouns for each image semantic center $\gamma$, and the number of nearest neighbors $\hat{N}$. The results are shown in Fig.~\ref{fig: parameter}.

\subsubsection{Expected compact cluster size $\tilde{N}$}
\label{sec: cluster size}
To see how the granularity of image semantics influences the final clustering performance, we test various choices of $\tilde{N}$ in Fig.~\ref{fig: compact_cluster_size}. As shown, TAC is stable across a reasonable range of $\tilde{N}$. However, since UCF-101 has an average cluster size much less than the default $\tilde{N}=300$, it encounters a performance drop on large cluster sizes due to overly coarse-grained semantics. Conversely, when the cluster size is overly small, the excessively fine-grained semantics leads to performance degradation on ImageNet-Dogs.

\subsubsection{Number of discriminative nouns $\gamma$}
To construct the text space, we classify all nouns into the image semantic centers and select the top $\gamma$ nouns of each center for retrieval. Here, we try various choices of $\gamma$ in Fig.~\ref{fig: number_of_discriminative_nouns}. As can be seen, a solitary noun is insufficient to cover the semantics of each image center. Conversely, an excessive number of nouns would falsely enrich the semantics, leading to inferior performance. Overall, TAC is stable across a typical range of discriminative noun number $\gamma$.

\subsubsection{Number of nearest neighbors $\hat{N}$}
To collaborate the text and image modalities, TAC mutually distills their neighborhood information. Here, we evaluate TAC with different numbers of nearest neighbors $\bar{N}$ in Fig.~\ref{fig: number_of_neighbors}. The results demonstrate that TAC is robust to diverse numbers of $\bar{N}$. Though a smaller choice of $\bar{N}$ leads to slight improvements on ImageNet-Dogs, we find the default $\bar{N}=50$ achieves stable results across different datasets.

\section{Conclusion}
In this paper, instead of focusing on exhaustive internal supervision signal construction, we innovatively propose leveraging the rich external knowledge, which has been regrettably overlooked before, to facilitate clustering. As a specific implementation, our TAC achieves state-of-the-art image clustering performance by leveraging textual semantics, demonstrating the effectiveness and promising prospect of the proposed externally guided clustering paradigm. In the future, the following directions could be worth exploring. On the one hand, in addition to the modalities this work focuses on, the external knowledge widely exists in different sources, domains, models, etc. For example, one could utilize the pre-trained object detection or semantic segmentation models to locate the semantic object for boosting image clustering. On the other hand, instead of focusing on image clustering, it is worth exploring the external knowledge for clustering other forms of data, such as text and point cloud. The challenges of the proposed externally guided clustering paradigm lie in \textit{i)}, choosing the appropriate external knowledge, and \textit{ii)}, effectively integrating the external knowledge to improve clustering. In practice, the selection and utilization of external knowledge should depend on the characteristics and prior knowledge about the data and task. Overall, we hope this work could serve as a catalyst, motivating more future studies on externally guided clustering, which is believable to be a promising direction for both methodology improvement and real-world application.

\section*{Acknowledgements}
This work was supported in part by NSFC under Grant 62176171,  U21B2040, 623B2075; in part by the Fundamental Research Funds for the Central Universities under Grant CJ202303; and in part by Sichuan Science and Technology Planning Project under Grant 24NSFTD0130.

%

\section*{Impact Statement}
This work proposes a new deep clustering paradigm by leveraging external knowledge. As a fundamental problem in machine learning, clustering has a wide range of applications, such as anomaly detection, person re-identification, community detection, etc. The proposed method is evaluated on public image datasets that are not at risk. However, just like any learning method, the performance of our method depends on data bias and cannot be guaranteed in more complex real-world applications. In this sense, it might bring some disturbances in decision-making and thus should be carefully used, especially in areas such as health care, autonomous vehicles, etc. Moreover, the proposed method requires manually setting the target cluster number. In real-world applications, one may resort to other cluster number estimation methods in the lack of cluster number prior.

\nocite{langley00}

\bibliography{reference}

\begin{thebibliography}{51}
\providecommand{\natexlab}[1]{#1}
\providecommand{\url}[1]{\texttt{#1}}
\expandafter\ifx\csname urlstyle\endcsname\relax
  \providecommand{\doi}[1]{doi: #1}\else
  \providecommand{\doi}{doi: \begingroup \urlstyle{rm}\Url}\fi

\bibitem[Cai et~al.(2009)Cai, He, Wang, Bao, and Han]{NMF}
Cai, D., He, X., Wang, X., Bao, H., and Han, J.
\newblock Locality preserving nonnegative matrix factorization.
\newblock In \emph{International Joint Conference on Artificial Intelligence}, volume~9, pp.\  1010--1015, 2009.

\bibitem[Cai et~al.(2023)Cai, Qiu, Chen, Zhang, and Chen]{SIC}
Cai, S., Qiu, L., Chen, X., Zhang, Q., and Chen, L.
\newblock Semantic-enhanced image clustering.
\newblock In \emph{AAAI Conference on Artificial Intelligence}, volume~37, pp.\  6869--6878, 2023.

\bibitem[Chang et~al.(2017{\natexlab{a}})Chang, Wang, Meng, Xiang, and Pan]{DAC}
Chang, J., Wang, L., Meng, G., Xiang, S., and Pan, C.
\newblock Deep adaptive image clustering.
\newblock In \emph{International Conference on Computer Vision}, pp.\  5879--5887, 2017{\natexlab{a}}.

\bibitem[Chang et~al.(2017{\natexlab{b}})Chang, Wang, Meng, Xiang, and Pan]{ImageNet_10_Dogs}
Chang, J., Wang, L., Meng, G., Xiang, S., and Pan, C.
\newblock Deep adaptive image clustering.
\newblock In \emph{International Conference on Computer Vision}, pp.\  5879--5887, 2017{\natexlab{b}}.

\bibitem[Chen et~al.(2020)Chen, Kornblith, Norouzi, and Hinton]{SimCLR}
Chen, T., Kornblith, S., Norouzi, M., and Hinton, G.
\newblock A simple framework for contrastive learning of visual representations.
\newblock In \emph{International Conference on Machine Learning}, pp.\  1597--1607. PMLR, 2020.

\bibitem[Cimpoi et~al.(2014)Cimpoi, Maji, Kokkinos, Mohamed, and Vedaldi]{DTD}
Cimpoi, M., Maji, S., Kokkinos, I., Mohamed, S., and Vedaldi, A.
\newblock Describing textures in the wild.
\newblock In \emph{IEEE Conference on Computer Vision and Pattern Recognition}, pp.\  3606--3613, 2014.

\bibitem[Coates et~al.(2011)Coates, Ng, and Lee]{STL}
Coates, A., Ng, A., and Lee, H.
\newblock An analysis of single-layer networks in unsupervised feature learning.
\newblock In \emph{International Conference on Artificial Intelligence and Statistics}, pp.\  215--223, 2011.

\bibitem[Dang et~al.(2021{\natexlab{a}})Dang, Deng, Yang, Wei, and Huang]{NNM}
Dang, Z., Deng, C., Yang, X., Wei, K., and Huang, H.
\newblock Nearest neighbor matching for deep clustering.
\newblock In \emph{IEEE/CVF Conference on Computer Vision and Pattern Recognition}, pp.\  13693--13702, 2021{\natexlab{a}}.

\bibitem[Dang et~al.(2021{\natexlab{b}})Dang, Deng, Yang, Wei, and Huang]{dang2021nearest}
Dang, Z., Deng, C., Yang, X., Wei, K., and Huang, H.
\newblock Nearest neighbor matching for deep clustering.
\newblock In \emph{IEEE/CVF Conference on Computer Vision and Pattern Recognition}, pp.\  13693--13702, 2021{\natexlab{b}}.

\bibitem[Deng et~al.(2009)Deng, Dong, Socher, Li, Li, and Fei-Fei]{ImageNet}
Deng, J., Dong, W., Socher, R., Li, L.-J., Li, K., and Fei-Fei, L.
\newblock Imagenet: A large-scale hierarchical image database.
\newblock In \emph{IEEE Conference on Computer Vision and Pattern Recognition}, pp.\  248--255. Ieee, 2009.

\bibitem[Dosovitskiy et~al.(2020)Dosovitskiy, Beyer, Kolesnikov, Weissenborn, Zhai, Unterthiner, Dehghani, Minderer, Heigold, Gelly, et~al.]{ViT}
Dosovitskiy, A., Beyer, L., Kolesnikov, A., Weissenborn, D., Zhai, X., Unterthiner, T., Dehghani, M., Minderer, M., Heigold, G., Gelly, S., et~al.
\newblock An image is worth 16x16 words: Transformers for image recognition at scale.
\newblock In \emph{International Conference on Learning Representations}, 2020.

\bibitem[Elhamifar \& Vidal(2013)Elhamifar and Vidal]{SSC}
Elhamifar, E. and Vidal, R.
\newblock Sparse subspace clustering: Algorithm, theory, and applications.
\newblock \emph{IEEE Transactions on Pattern Analysis and Machine Intelligence}, 35\penalty0 (11):\penalty0 2765--2781, 2013.

\bibitem[Ester et~al.(1996)Ester, Kriegel, Sander, Xu, et~al.]{DBSCAN}
Ester, M., Kriegel, H.-P., Sander, J., Xu, X., et~al.
\newblock A density-based algorithm for discovering clusters in large spatial databases with noise.
\newblock In \emph{KDD}, volume~96, pp.\  226--231, 1996.

\bibitem[Ghasedi~Dizaji et~al.(2017)Ghasedi~Dizaji, Herandi, Deng, Cai, and Huang]{DeepClusteringGhasedi}
Ghasedi~Dizaji, K., Herandi, A., Deng, C., Cai, W., and Huang, H.
\newblock Deep clustering via joint convolutional autoencoder embedding and relative entropy minimization.
\newblock In \emph{International Conference on Computer Vision}, pp.\  5736--5745, 2017.

\bibitem[Gowda \& Krishna(1978)Gowda and Krishna]{AgglomerativeClustering}
Gowda, K.~C. and Krishna, G.
\newblock Agglomerative clustering using the concept of mutual nearest neighbourhood.
\newblock \emph{Pattern Recognition}, 10\penalty0 (2):\penalty0 105--112, 1978.

\bibitem[Grill et~al.(2020)Grill, Strub, Altch{\'e}, Tallec, Richemond, Buchatskaya, Doersch, Avila~Pires, Guo, Gheshlaghi~Azar, et~al.]{BYOL}
Grill, J.-B., Strub, F., Altch{\'e}, F., Tallec, C., Richemond, P., Buchatskaya, E., Doersch, C., Avila~Pires, B., Guo, Z., Gheshlaghi~Azar, M., et~al.
\newblock Bootstrap your own latent-a new approach to self-supervised learning.
\newblock \emph{Advances in Neural Information Processing Systems}, 33:\penalty0 21271--21284, 2020.

\bibitem[Guo et~al.(2017)Guo, Liu, Zhu, and Yin]{DeepClusteringGuo}
Guo, X., Liu, X., Zhu, E., and Yin, J.
\newblock Deep clustering with convolutional autoencoders.
\newblock In \emph{International Conference on Neural Information Processing}, pp.\  373--382. Springer, 2017.

\bibitem[He et~al.(2020)He, Fan, Wu, Xie, and Girshick]{MOCO}
He, K., Fan, H., Wu, Y., Xie, S., and Girshick, R.
\newblock Momentum contrast for unsupervised visual representation learning.
\newblock In \emph{IEEE/CVF Conference on Computer Vision and Pattern Recognition}, pp.\  9729--9738, 2020.

\bibitem[Hu et~al.(2017)Hu, Miyato, Tokui, Matsumoto, and Sugiyama]{IMSAT}
Hu, W., Miyato, T., Tokui, S., Matsumoto, E., and Sugiyama, M.
\newblock Learning discrete representations via information maximizing self-augmented training.
\newblock In \emph{International Conference on Machine Learning}, pp.\  1558--1567. PMLR, 2017.

\bibitem[Huang et~al.(2020)Huang, Gong, and Zhu]{PICA}
Huang, J., Gong, S., and Zhu, X.
\newblock Deep semantic clustering by partition confidence maximisation.
\newblock In \emph{IEEE/CVF Conference on Computer Vision and Pattern Recognition}, June 2020.

\bibitem[Huang et~al.(2022)Huang, Chen, Zhang, and Shan]{ProPos}
Huang, Z., Chen, J., Zhang, J., and Shan, H.
\newblock Learning representation for clustering via prototype scattering and positive sampling.
\newblock \emph{IEEE Transactions on Pattern Analysis and Machine Intelligence}, 2022.

\bibitem[Ji et~al.(2019)Ji, Henriques, and Vedaldi]{IIC}
Ji, X., Henriques, J.~F., and Vedaldi, A.
\newblock Invariant information clustering for unsupervised image classification and segmentation.
\newblock In \emph{International Conference on Computer Vision}, pp.\  9865--9874, 2019.

\bibitem[Krizhevsky \& Hinton(2009)Krizhevsky and Hinton]{CIFAR}
Krizhevsky, A. and Hinton, G.
\newblock Learning multiple layers of features from tiny images.
\newblock \emph{Master's thesis, Department of Computer Science, University of Toronto}, 2009.

\bibitem[Li et~al.(2022)Li, Li, Xiong, and Hoi]{BLIP}
Li, J., Li, D., Xiong, C., and Hoi, S.
\newblock Blip: Bootstrapping language-image pre-training for unified vision-language understanding and generation.
\newblock In \emph{International Conference on Machine Learning}, pp.\  12888--12900. PMLR, 2022.

\bibitem[Li et~al.(2021{\natexlab{a}})Li, Gao, Niu, Xiao, Liu, Liu, Wu, and Wang]{UNIMO}
Li, W., Gao, C., Niu, G., Xiao, X., Liu, H., Liu, J., Wu, H., and Wang, H.
\newblock Unimo: Towards unified-modal understanding and generation via cross-modal contrastive learning.
\newblock In \emph{Annual Meeting of the Association for Computational Linguistics}, pp.\  2592--2607, 2021{\natexlab{a}}.

\bibitem[{Li} et~al.(2020){Li}, {Zhang}, {Wang}, and {Zhang}]{deepclustering_wangqi}
{Li}, X., {Zhang}, R., {Wang}, Q., and {Zhang}, H.
\newblock Autoencoder constrained clustering with adaptive neighbors.
\newblock \emph{IEEE Transactions on Neural Networks and Learning Systems}, pp.\  1--7, 2020.

\bibitem[Li et~al.(2021{\natexlab{b}})Li, Hu, Liu, Peng, Zhou, and Peng]{CC}
Li, Y., Hu, P., Liu, Z., Peng, D., Zhou, J.~T., and Peng, X.
\newblock Contrastive clustering.
\newblock In \emph{AAAI Conference on Artificial Intelligence}, volume~35, pp.\  8547--8555, 2021{\natexlab{b}}.

\bibitem[Liu et~al.(2012)Liu, Lin, Yan, Sun, Yu, and Ma]{LRR}
Liu, G., Lin, Z., Yan, S., Sun, J., Yu, Y., and Ma, Y.
\newblock Robust recovery of subspace structures by low-rank representation.
\newblock \emph{IEEE Transactions on Pattern Analysis and Machine Intelligence}, 35\penalty0 (1):\penalty0 171--184, 2012.

\bibitem[Liu et~al.(2017)Liu, Shen, and Tsang]{liu2017sparse}
Liu, W., Shen, X., and Tsang, I.
\newblock Sparse embedded k-means clustering.
\newblock In \emph{Advances in Neural Information Processing Systems}, pp.\  3319--3327, 2017.

\bibitem[MacQueen et~al.(1967)]{Kmeans}
MacQueen, J. et~al.
\newblock Some methods for classification and analysis of multivariate observations.
\newblock In \emph{Proceedings of the fifth Berkeley symposium on mathematical statistics and probability}, volume~1, pp.\  281--297. Oakland, CA, USA, 1967.

\bibitem[Miller(1995)]{WordNet}
Miller, G.~A.
\newblock Wordnet: a lexical database for english.
\newblock \emph{Communications of the ACM}, 38\penalty0 (11):\penalty0 39--41, 1995.

\bibitem[Nie et~al.(2011)Nie, Zeng, Tsang, Xu, and Zhang]{SpectralClusteringNie}
Nie, F., Zeng, Z., Tsang, I.~W., Xu, D., and Zhang, C.
\newblock Spectral embedded clustering: A framework for in-sample and out-of-sample spectral clustering.
\newblock \emph{IEEE Transactions on Neural Networks}, 22\penalty0 (11):\penalty0 1796--1808, 2011.

\bibitem[Nie et~al.(2016)Nie, Wang, Jordan, and Huang]{nie2016constrained}
Nie, F., Wang, X., Jordan, M.~I., and Huang, H.
\newblock The constrained laplacian rank algorithm for graph-based clustering.
\newblock In \emph{AAAI Conference on Artificial Intelligence}, pp.\  1969--1976. Citeseer, 2016.

\bibitem[Niu et~al.(2022)Niu, Shan, and Wang]{SPICE}
Niu, C., Shan, H., and Wang, G.
\newblock Spice: Semantic pseudo-labeling for image clustering.
\newblock \emph{IEEE Transactions on Image Processing}, 31:\penalty0 7264--7278, 2022.

\bibitem[Peng et~al.(2016)Peng, Xiao, Feng, Yau, and Yi]{DeepClusteringPeng2016}
Peng, X., Xiao, S., Feng, J., Yau, W.-Y., and Yi, Z.
\newblock Deep subspace clustering with sparsity prior.
\newblock In \emph{International Joint Conference on Artificial Intelligence}, pp.\  1925--1931, 2016.

\bibitem[Peng et~al.(2018)Peng, Feng, Xiao, Yau, Zhou, and Yang]{DeepClusteringPeng2018}
Peng, X., Feng, J., Xiao, S., Yau, W.-Y., Zhou, J.~T., and Yang, S.
\newblock Structured autoencoders for subspace clustering.
\newblock \emph{IEEE Transactions on Image Processing}, 27\penalty0 (10):\penalty0 5076--5086, 2018.

\bibitem[Radford et~al.(2021)Radford, Kim, Hallacy, Ramesh, Goh, Agarwal, Sastry, Askell, Mishkin, Clark, et~al.]{CLIP}
Radford, A., Kim, J.~W., Hallacy, C., Ramesh, A., Goh, G., Agarwal, S., Sastry, G., Askell, A., Mishkin, P., Clark, J., et~al.
\newblock Learning transferable visual models from natural language supervision.
\newblock In \emph{International Conference on Machine Learning}, pp.\  8748--8763. PMLR, 2021.

\bibitem[Shen et~al.(2021)Shen, Shen, Wang, Qin, Torr, and Shao]{TCC}
Shen, Y., Shen, Z., Wang, M., Qin, J., Torr, P., and Shao, L.
\newblock You never cluster alone.
\newblock \emph{Advances in Neural Information Processing Systems}, 34:\penalty0 27734--27746, 2021.

\bibitem[Soomro et~al.(2012)Soomro, Zamir, and Shah]{UCF101}
Soomro, K., Zamir, A.~R., and Shah, M.
\newblock Ucf101: A dataset of 101 human actions classes from videos in the wild.
\newblock \emph{arXiv preprint arXiv:1212.0402}, 2012.

\bibitem[Tao et~al.(2020)Tao, Takagi, and Nakata]{IDFD}
Tao, Y., Takagi, K., and Nakata, K.
\newblock Clustering-friendly representation learning via instance discrimination and feature decorrelation.
\newblock In \emph{International Conference on Learning Representations}, 2020.

\bibitem[Tsai et~al.(2020)Tsai, Li, and Zhu]{MiCE}
Tsai, T.~W., Li, C., and Zhu, J.
\newblock Mice: Mixture of contrastive experts for unsupervised image clustering.
\newblock In \emph{International Conference on Learning Representations}, 2020.

\bibitem[Van et~al.(2020)Van, Vandenhende, Georgoulis, Proesmans, and Van~Gool]{SCAN}
Van, G.~W., Vandenhende, S., Georgoulis, S., Proesmans, M., and Van~Gool, L.
\newblock Scan: Learning to classify images without labels.
\newblock In \emph{European Conference on Computer Vision}, pp.\  268--285. Springer, 2020.

\bibitem[Vaswani et~al.(2017)Vaswani, Shazeer, Parmar, Uszkoreit, Jones, Gomez, Kaiser, and Polosukhin]{Transformer}
Vaswani, A., Shazeer, N., Parmar, N., Uszkoreit, J., Jones, L., Gomez, A.~N., Kaiser, {\L}., and Polosukhin, I.
\newblock Attention is all you need.
\newblock \emph{Advances in Neural Information Processing Systems}, 30, 2017.

\bibitem[Wang et~al.(2022)Wang, Bao, Dong, Bjorck, Peng, Liu, Aggarwal, Mohammed, Singhal, Som, et~al.]{Beit3}
Wang, W., Bao, H., Dong, L., Bjorck, J., Peng, Z., Liu, Q., Aggarwal, K., Mohammed, O.~K., Singhal, S., Som, S., et~al.
\newblock Image as a foreign language: Beit pretraining for all vision and vision-language tasks.
\newblock \emph{arXiv preprint arXiv:2208.10442}, 2022.

\bibitem[Wang et~al.(2020)Wang, Li, Wang, Nie, and Li]{wang2020large}
Wang, Z., Li, Z., Wang, R., Nie, F., and Li, X.
\newblock Large graph clustering with simultaneous spectral embedding and discretization.
\newblock \emph{IEEE Transactions on Pattern Analysis and Machine Intelligence}, 2020.

\bibitem[Wu et~al.(2019)Wu, Long, Wang, Qian, Li, Lin, and Zha]{DCCM}
Wu, J., Long, K., Wang, F., Qian, C., Li, C., Lin, Z., and Zha, H.
\newblock Deep comprehensive correlation mining for image clustering.
\newblock In \emph{International Conference on Computer Vision}, pp.\  8150--8159, 2019.

\bibitem[Xie et~al.(2016)Xie, Girshick, and Farhadi]{DEC}
Xie, J., Girshick, R., and Farhadi, A.
\newblock Unsupervised deep embedding for clustering analysis.
\newblock In \emph{International Conference on Machine Learning}, pp.\  478--487, 2016.

\bibitem[Yang et~al.(2016)Yang, Parikh, and Batra]{JULE}
Yang, J., Parikh, D., and Batra, D.
\newblock Joint unsupervised learning of deep representations and image clusters.
\newblock In \emph{IEEE Conference on Computer Vision and Pattern Recognition}, pp.\  5147--5156, 2016.

\bibitem[Zelnik-Manor \& Perona(2005)Zelnik-Manor and Perona]{SpectralClustering}
Zelnik-Manor, L. and Perona, P.
\newblock Self-tuning spectral clustering.
\newblock In \emph{Advances in Neural Information Processing Systems}, pp.\  1601--1608, 2005.

\bibitem[Zhong et~al.(2021)Zhong, Wu, Chen, Huang, Deng, Nie, Lin, and Hua]{GCC}
Zhong, H., Wu, J., Chen, C., Huang, J., Deng, M., Nie, L., Lin, Z., and Hua, X.-S.
\newblock Graph contrastive clustering.
\newblock In \emph{International Conference on Computer Vision}, pp.\  9224--9233, 2021.

\bibitem[Zhou et~al.(2022)Zhou, Loy, and Dai]{DenseCLIP}
Zhou, C., Loy, C.~C., and Dai, B.
\newblock Extract free dense labels from clip.
\newblock In \emph{European Conference on Computer Vision}, pp.\  696--712. Springer, 2022.

\end{thebibliography}
\bibliographystyle{icml2024}

\newpage
\appendix
\onecolumn

\section{Variants of text counterpart construction.}
\label{apx: text construct}

\begin{table}[h]
\centering
\caption{Clustering performance of TAC using different clustering methods for text counterpart construction. (AC: agglomerative clustering, SC: spectral clustering, $r$: resolution of Louvain clustering, None: using all nouns from WordNet)}
\label{tab: noun filter}
\resizebox{0.55\columnwidth}{!}{%
\begin{tabular}{@{}cccccccc@{}}
\toprule
\multirow{2}{*}{Method} &
  \multirow{2}{*}{\begin{tabular}[c]{@{}c@{}}Semantic\\ Space\end{tabular}} &
  \multicolumn{3}{c}{ImageNet-Dogs} &
  \multicolumn{3}{c}{UCF-101} \\ \cmidrule(l){3-8} 
                      &         & NMI & ACC & ARI & NMI & ACC & ARI \\ \midrule
\multirow{8}{*}{\begin{tabular}[c]{@{}c@{}}TAC\\ (no train)\end{tabular}} &
  k-means & 75.1 & 75.1 & 63.6 & 81.6 & 61.3 & 52.4 \\
                      & AC      & 73.4 & 72.0 & 61.2 & \underline{81.9} & \underline{63.7} & \textbf{54.8} \\
                      & SC      & \textbf{77.4} & \underline{75.2} & \underline{65.9} & \textbf{82.2} & \textbf{65.5} & \underline{54.7} \\
                      & DBSCAN  & 68.8 & 64.3 & 51.0 & 81.1 & 61.8 & 52.3 \\
                      & {Louvain ($r$=1)}  & {77.0} & {75.1} & {65.0} & {78.6} & {58.3} & {47.9} \\
                      & {Louvain ($r$=5)}  & {\underline{77.1}} & {\textbf{78.3}} & {\textbf{66.9}} & {81.3} & {61.7} & {54.0} \\
                      & {Louvain ($r$=10)}  & {75.7} & {72.9} & {62.7} & {80.9} & {60.8} & {52.5} \\
                      & None    & 70.3 & 68.7 & 53.6 & 81.3 & 63.2 & 52.8 \\ \midrule 
\multirow{8}{*}{TAC} & k-means & \textbf{80.6} & 83.0 & \underline{72.2} & \underline{82.3} & 68.7 & 60.1 \\
                      & AC      & 78.4 & 81.7 & 69.5 & \textbf{82.4} & \textbf{69.3} & \underline{60.2} \\
                      & SC      & 79.2 & 83.1 & 70.8 & \underline{82.3} & \underline{69.1} & \textbf{60.4} \\ 
                      & DBSCAN  & 75.5 & 80.4 & 65.5 & 80.6 & 66.2 & 56.4 \\
                      & {Louvain ($r$=1)}  & {78.5} & {83.6} & {70.0} & {78.3} & {61.7} & {53.0} \\
                      & {Louvain ($r$=5)}  & {\underline{79.8}} & {\textbf{85.6}} & {\textbf{72.6}} & {81.7} & {68.3} & {59.1} \\
                      & {Louvain ($r$=10)}  & {79.6} & {\underline{85.1}} & {71.9} & {82.1} & {68.1} & {59.4} \\
                      & None    & 75.7 & 78.7 & 66.0 & 81.2 & 67.0 & 58.3 \\ \bottomrule
\end{tabular}%
}
\end{table}

Recall that to select representation nouns for text counterpart construction, we first classify all nouns from WordNet to image semantic centers found by applying k-means on image embeddings. Here, we investigate the robustness and necessity of the noun selection step. Specifically, we adopt three other classic clustering methods to compute semantic centers, including agglomerative clustering (AC), spectral clustering (SC), and DBSCAN. For AC and SC, we set the target cluster number to the same as k-means. For DBSCAN, we tune the density parameter until it produces the same number of clusters. As shown in Table~\ref{tab: noun filter}, the training-free TAC achieves better performance with SC, while the performance is similar among k-means, AC, and SC when further boosted with cross-modal mutual distillation. The performance degradation on DBSCAN is probably due to the poor quality of image embeddings. In practice, we find DBSCAN tends to treat a portion of samples as outliers, and thus it cannot precisely cover the image semantics, leading to suboptimal performance. Moreover, we test the Louvain clustering algorithm that could estimate the cluster number given the cluster resolution. One could see that Louvain clustering gives promising results with an appropriate choice of resolution. Nevertheless, almost all cluster number estimation methods require manually setting a granularity parameter like the resolution here in Louvain. Such a process is similar to our simple estimation based on the sample size. To investigate the necessity to filter discriminative nouns, we further append a baseline by retrieving text counterparts from all nouns. According to the results, TAC encounters a performance drop on both datasets, but the influence is milder on UCF-101, which could be attributed to the richer image semantics in that dataset. In summary, the results demonstrate the effectiveness of discriminative noun selection, as well as the robustness of TAC against different clustering methods used for text counterpart construction.

\section{The textual semantic space constructed by TAC.}
\label{apx: nouns}
To provide an intuitive understanding of the textual space constructed by TAC, we provide the discriminative nouns selected for all datasets. Due to the space limitation, only the thirty most discriminative nouns are shown. As can be seen, for object clustering datasets including STL, CIFAR, and ImageNet, the selected discriminative nouns directly match the name of objects. The results are more intriguing on the DTD and UCF-101 datasets with textures and actions as the clustering criterion, respectively. For the DTD dataset, some nouns directly match the adjectives that describe textures. For example, "Belgian waffle" matches the "waffled" texture, and "honeycomb" matches the "honeycombed" texture. For other selected nouns that do not have a direct matching, they turn out to have close relationships with those textures. For example, the nouns "garden lettuce", "peony", and "Peruvian lily" correspond to the "frilly" texture, "grevy's zebra" corresponds to the "striped" texture, and "chessboard" corresponds to the "chequered" texture. In other words, these nouns describe or reflect the texture and can thus benefit the discrimination between images of different textures. For the UCF-101 dataset, most selected nouns correspond to the object that actions interact with. For example, the "snooker table" in "billiard hall" is used for "Billiards" and the "typewriter keyboard" is used for "Typing". There also exist some gerundial nouns that directly refer to the actions such as "cliff diving" and "touch typing". The close connection between the nouns and actions explains the performance improvement in image clustering. To summarize, external nouns from WordNet would be closely related to the semantics in images, either directly or indirectly. As a result, these nouns could provide more compact semantics and benefit clustering.

\begin{table}[h]
\centering
\caption{Top-30 selected discriminative nouns for image semantic centers from different datasets.}
\label{tab: nouns}
\resizebox{\columnwidth}{!}{%
\begin{tabular}{@{}lp{15cm}@{}}
\toprule
Dataset       & Selected Discriminative Nouns \\ \midrule
STL-10        & floatplane, titi monkey, sand cat, whitetail deer, harness horse, black billed cuckoo, garbage truck, fire truck, Lipizzan, container ship, ocean liner, trucking rig, aerobatics, Angora cat, airline, black and tan terrier, electric automobile \\ \midrule
CIFAR-10      & spadefoot toad, field sparrow, curassow, Lipizzan, sable antelope, black fronted bush shrike, chameleon tree frog, whitetail deer, fire truck, waterbuck, emu, hartebeest, dressage, containership, pratincole, woodland caribou, hydroplane racing, banana boat, yacht race, yellow breasted chat, pen tailed tree shrew, elk, wagtail, stonefish, trucking rig, clipper ship, Texas horned lizard, stealth bomber, articulated lorry, fire department \\ \midrule
CIFAR-20      & Iceland poppy, Lepiota procera, oceanic whitetip shark, prairie sunflower, goblet, common dolphin, carabid beetle, sunflower, rosy boa, trolleybus, soda can, school bus, Peromyscus maniculatus, characin fish, banded gecko, Arabian camel, armoured combat vehicle, diesel electric locomotive, African elephant, bunk bed, tandem bicycle, sandbar shark, common wallaby, European spider crab, eastern chimpanzee, navel orange, edmontosaurus, Kodiak bear, lawn mower, tractor \\ \midrule
ImageNet-10   & crested penguin, snow leopard, Graf Zeppelin, navel orange, Maltese terrier, soccer ball, blimp, candied citrus peel, airship, serval, dirigible, containership, articulated lorry, trailer truck, tractor trailer, sports car, tufted puffin, soccer player, wire haired terrier, roadster, Sealyham terrier, soft coated wheaten terrier, sporting dog, airline, turboprop, flying bomb, wind bell, fruit tree, Antarctic Peninsula, airliner \\ \midrule
ImageNet-Dogs & Doberman pinscher, giant schnauzer, Norwegian elkhound, clumber spaniel, chowchow, Shetland sheepdog, Welsh springer spaniel, pug, standard schnauzer, schipperke, basset, beach, merino sheep, Arctic wolf, keeshond, Maltese terrier, standard poodle, snowfall, swimming hole, chromolithography, sleeping partner, chipping sparrow, dog show, Persian cat, Pomeranian, golden retriever, triple jump, meadow jumping mouse, fieldwork, harness race \\ \midrule
DTD           & butterflyfish, garden lettuce, peony, grevy's zebra, chessboard, Peruvian lily, pothole, Belgian waffle, turban squash, African elephant, anchor rope, chainlink fence, wicker basket, sweetsop, honeycomb, komondor, cytologic smear, rood screen, orb weaving spider, pillow lace, fluorite, proboscis monkey, birch bark, grape leaf begonia, sunset, zebrawood, stockinette stitch, lecanora, houndstooth check, black crappie \\ \midrule
UCF-101       & sitar, cliff diving, snooker table, blackboard, typewriter keyboard, billiard hall, marching band, darning needle, touch typing, Islamic Army of Aden, piano sonata, violoncellist, table tennis, koto player, contradance, bowling alley, Seattle Slew, tai chi chuan, parade, sumo ring, candlepin bowling, pelican crossing, cymbalist, Armenian Secret Army for the Liberation of Armenia, tenor drum, woodwind instrument, Panjabi, Victory Day, cyclist, Belmont Stakes \\ \midrule
ImageNet-1K   & Cypripedium fasciculatum, fireboat, swamphen, colobus monkey, plaque, komondor, limpkin, chasuble, European black grouse, nine banded armadillo, sulphur crested cockatoo, ruffed grouse, common stinkhorn, genus Cypripedium, Geastrum coronatum, ptarmigan, red breasted merganser, tobacconist shop, oystercatcher, axolotl, slate colored junco, purple gallinule, black capped chickadee, Tibetan mastiff, redshank, red legged partridge, Polaroid camera, pygmy marmoset, cherimoya, sharp tailed grouse \\ \bottomrule
\end{tabular}%
}
\end{table}

%

\end{document}